\documentclass{article}

\usepackage{graphicx}
\usepackage{amsmath}
\usepackage{amssymb}
\usepackage{booktabs}
\usepackage{makecell}
\usepackage{float}
\usepackage{svg}
\usepackage{multirow}

\title{Comparative Evaluation of CNN Architectures for Neural Style Transfer in Indonesian Batik Motif Generation: A Comprehensive Study}

\author{
Happy Gery Pangestu$^{1}$,
Andi Prademon Yunus$^{2,3}$,
Siti Khomsah$^{1}$\\[0.5em]
\small
$^{1}$Department of Data Science, Telkom University, Purwokerto Campus, Indonesia\\
\small$^{2}$Department of Informatics Engineering, Telkom University, Purwokerto Campus, Indonesia\\
\small$^{3}$Center of Excellece Human Centric Engineering, Telkom University, Indonesia\\
\texttt{andiay@telkomuniversity.ac.id}
}

\date{}

\begin{document}
\maketitle

\begin{abstract}
Neural Style Transfer (NST) provides a computational framework for the digital preservation and generative exploration of Indonesian batik motifs; however, existing approaches remain largely centered on VGG-based architectures whose strong stylistic expressiveness comes at the cost of high computational and memory demands, that limits practical deployment in resource-limited environments. This study presents a systematic comparative analysis of five widely used CNN backbones, namely VGG16, VGG19, Inception~V3, ResNet50, and ResNet101, based on 245 controlled experiments combining quantitative metrics, qualitative assessment, and statistical analysis to examine the trade-off between structural preservation, stylistic behavior, and computational efficiency. The results show that backbone selection does not yield statistically significant differences in structural similarity, as confirmed by ANOVA on SSIM ($p = 0.83$), indicating comparable levels of structural preservation rather than equivalent stylistic quality. Within this context, ResNet-based architectures achieve approximately $5$--$6\times$ faster convergence than VGG models while maintaining similar perceptual similarity (LPIPS $\approx 0.53$) and requiring over $16\times$ fewer FLOPs (0.63 vs.\ 10.12 GFLOPs). Qualitative analysis reveals consistent stylistic trade-offs, with VGG producing denser painterly textures, ResNet favoring geometric stability and canting stroke preservation with milder stylization, and Inception~V3 exhibiting intermediate but noisier behavior. These findings reposition architectural choice in NST from maximizing stylistic intensity toward efficiency-aware and structure-preserving deployment, highlighting ResNet-based backbones as a practical foundation for scalable, industry-oriented batik generation.
\end{abstract}

\section{Introduction}

In light of the evolving demands for both safeguarding and reimagining cultural heritage, particularly in the context of Indonesian batik \cite{Ratnawati_2022} \cite{kusumasari2019business} \cite{chu2025modernizing}, this study explores the role of Generative Artificial Intelligence as a vehicle for cultural expression. Traditional textile practices such as batik encapsulate deeply embedded visual grammars, layered with symbolic, structural, and regional diversity, that present both a challenge and an opportunity for algorithmic modeling. While digital tools have long been employed in heritage conservation, the advent of generative models introduces a shift in paradigm, from passive documentation to active co-creation.

This research specifically investigates Neural Style Transfer (NST) as a computational technique for generating new batik-inspired motifs. Originally proposed by Gatys et al. \cite{gatys2015neural}, NST operates on the principle of disentangling visual content from artistic style using the representational power of deep convolutional neural networks. 

The core mechanism of NST relies on a pre-trained Convolutional Neural Network (CNN), typically one trained for large-scale image classification like VGG \cite{10578183}. Within this network, the activations of deeper layers, which capture high level object features, are used to represent the image's content. Conversely, the correlations between feature activations in the network's shallower layers, which encode textural and pattern information, are used to represent the style \cite{8578939}. By optimizing a new image to match the content representation of one source image and the style representation of another, NST can generate a novel artistic fusion \cite{london}.

Since it was first proposed, the field of generative art has evolved rapidly. The original optimization-based NST algorithm, while foundational, has been succeeded by faster feed-forward networks capable of applying learned styles in real time, and more recently by large-scale diffusion models such as Stable Diffusion and DALL·E, which offer unprecedented levels of photorealism and semantic control through text-based prompts. Despite these advances, a deep understanding of the principles established by classical NST remains essential \cite{jing2020neural}, as the explicit manipulation of deep feature spaces to disentangle and recombine content and style underpins many contemporary generative frameworks \cite{CAI2023108723}. Notably, while diffusion-based models represent the current frontier in generative synthesis, their stochastic generation process can introduce structural hallucinations, that leads to unintended distortions of canonical geometric patterns. Recent advances in structure-guided diffusion, such as ControlNet and related conditional adapters \cite{zhang2023addingconditionalcontroltexttoimage}, have substantially alleviated this issue by explicitly constraining the generative process using auxiliary structural inputs, which allows high-fidelity preservation of geometric layouts and line structures. Such behavior poses a significant limitation for cultural heritage applications, where strict preservation of structural integrity is critical. In the context of Indonesian batik, whose motifs encode symbolic and philosophical meanings through precise geometric arrangements, even minor structural deviations may compromise cultural authenticity. However, despite their representational strengths, structure-guided diffusion models typically require high-end computational resources (e.g., $\geq$8--6.8 GB VRAM)\cite{li2024controlnetimprovingconditionalcontrols}, which restricts their practical deployment in resource-constrained production environments. In contrast, classical optimization-based NST operates in a deterministic manner, explicitly preserving the spatial structure of the content image while modulating only its textural and stylistic characteristics. This property makes classical NST particularly suitable for heritage sensitive visual computing tasks, where the faithful retention of original line structures, such as hand-drawn canting strokes is essential. Consequently, this study deliberately focuses on the classical NST framework to conduct a \textit{first-principles} analysis, providing a rigorous and detailed investigation into how its core technical parameters interact with the unique and complex visual language of Indonesian batik.

This paper presents a systematic technical investigation into the application of classical Neural Style Transfer for the generation of novel batik motifs. Building upon our preliminary work \cite{pangestu2025batik}, this research refines the focus, updates the technical context with contemporary literature, and provides a deeper analysis of the underlying mechanisms. The primary objective is to demystify the NST process in the context of a specific cultural art form, providing a practical framework for its use as a creative tool. We believe this study strongly  contributes:
\begin{enumerate}
  \item A comprehensive comparative evaluation of five pre-trained CNN architectures (VGG16, VGG19, Inception V3, ResNet50, ResNet101) for batik style transfer, using rigorous statistical analysis to assess quality differences and identify key performance factors.
  \item Empirical demonstration that architectural choices yield comparable output quality, whereas training efficiency constitutes the dominant differentiating factor, offering practical guidance for balancing computational cost and visual fidelity.
  \item Systematic ablation study examining hyperparameter sensitivity across content/style weight ratios, feature layer selections, and learning rates, revealing that hyperparameter tuning provides substantially greater control over visual characteristics than architectural selection.
  \item A visual quality assessment revealing distinct architectural characteristics in terms of stylization patterns and texture handling, emphasizing that both architectural and hyperparameter selection should prioritize aesthetic preferences and computational constraints rather than quantitative metric rankings alone.
\end{enumerate}

By addressing these objectives, this research aims to provide valuable insights for computer scientists interested in generative models, as well as for artists and cultural practitioners who seek to engage with AI as a collaborative partner in the evolution of traditional art forms.

\subsection{Related Work}
Neural style transfer was first introduced by Gatys \textit{et al.} \cite{gatys2016image}, who employed a pretrained VGG network to disentangle and recombine image content and style. In this formulation, content loss is computed from high-level feature activations, where the style loss relies on gram matrices to represent correlations between feature maps at different layers. Building on this formulation, later studies have aimed to improve computational efficiency as well as practical usability.: Johnson \textit{et al.} \cite{johnson2016perceptual} and Ulyanov \textit{et al.} \cite{ulyanov2016texture} proposed feed-forward networks for real-time stylization, while Zhang \textit{et al.} \cite{zhang2017stylebank} introduced a StyleBank framework to support multiple styles within a single model. Despite these advances, the continued reliance on VGG-based feature extractors in NST reflects a methodological inertia, as their dominance has rarely been questioned in terms of computational efficiency or suitability for resource-constrained deployment.

Our preliminary exploration~\cite{pangestu2025batik} investigated hyperparameter sensitivity and VGG feature dynamics for batik motif generation, with a particular focus on pooling strategies and layer selection and their influence on cultural pattern coherence. While this study provided foundational insights into parameter interactions within VGG-based NST, it was limited to a single architectural family. The present work substantially extends this foundation by (i) systematically comparing five distinct CNN backbones, such as VGG16, VGG19, Inception~V3, ResNet50, and ResNet101, rather than relying solely on VGG variants; (ii) conducting rigorous statistical analysis across 245 controlled experiments to quantify architectural effects; and (iii) performing comprehensive ablation studies to examine hyperparameter sensitivity across heterogeneous network designs.

Beyond methodological curiosity, the motivation for evaluating alternative architectures is grounded in practical deployment considerations. Classical VGG networks are computationally expensive and memory-intensive\cite{sherif}, which limits their applicability in low-resource or edge-computing scenarios. In the context of Indonesian batik production, which is predominantly carried out by small and medium-sized enterprises (SMEs), access to high-end computing infrastructure is often unavailable. From this perspective, exploring lighter or more parameter-efficient architectures such as ResNet and Inception is not merely an optimization exercise, but a necessary step toward assessing the feasibility of neural style transfer on consumer-grade hardware. Importantly, such feasibility is closely tied to the ability of pretrained models to transfer robust visual representations across domains without extensive retraining.

Transfer learning has been widely adopted to adapt pretrained CNNs to new visual domains \cite{yosinski2014transferable}, and architectures such as ResNet \cite{he2016deep} and Inception \cite{szegedy2016rethinking} have demonstrated superior representational efficiency in large-scale image recognition tasks. However, their implications for style transfer, particularly for structurally constrained, culturally significant patterns such as batik remain underexplored. To our knowledge, no prior study has systematically examined how architectural choice interacts with NST hyperparameters in the context of batik motif generation. This gap motivates our comparative analysis of network backbones and training configurations, with an emphasis on interpretability, robustness, and practical deployability.

The selection of CNN architecture as a feature extractor plays a decisive role in determining the final outcome of style transfer \cite{khan2021image}.  While many early and subsequent works have defaulted to using VGG networks  \cite{9918891}, comparative studies have revealed important nuances. Empirical evaluations consistently show that VGG networks, particularly those without batch normalization layers, often yield the best performance for tasks based on perceptual loss. This is attributed to VGG's simple, sequential architecture of stacked 3x3 filters, which creates a clean, uniform feature hierarchy well-suited for the optimization process \cite{pihlgren2024systematicperformanceanalysisdeep}. 
\section{Methodology}
Our experimental investigation was conducted in two phases: a main batch evaluation and a supplementary ablation study. The main batch evaluation comprised a total of 145 experiments across five CNN architectures. Each architecture (VGG16, VGG19, Inception~V3, ResNet50, and ResNet101) processed 49 distinct content--style image pairs, which produced a balanced experimental design that enabled statistically meaningful architectural comparisons while remaining computationally tractable. The selected image pairs were curated to represent diverse batik motif characteristics, covering varying levels of geometric complexity, motif density, and structural regularity, thereby ensuring coverage of both simple and highly intricate pattern compositions.

The supplementary ablation study consisted of 49 additional experiments aimed at analyzing hyperparameter sensitivity. Three key hyperparameters were systematically varied: (i) content--style weight ratios, (ii) feature extraction layer selections, and (iii) learning rates. The ablation experiments were organized into three separate sets, each isolating a single hyperparameter while fixing the remaining parameters at their baseline values. This controlled factorial design allowed us to assess the individual influence of each hyperparameter on style transfer quality, convergence behavior, and structural stability.

\subsection{Style Transfer Method}
Style transfer is used for transferring the style from the second batik to the first batik. Firstly, the content and style images are preprocessed by rescaling them to the same size and normalizing them to match the ImageNet data distribution. This preprocessing ensures that the images have the same size and distribution as those used to train the ImageNet CNN models, which are employed in this research. Secondly, the trained CNN models employed to extract features from each content and style images. For the result image, the content image is used as a baseline instead of initializing from a random latent space for training efficiency. Then, for the style images, a style representation is constructed using correlations between the different filter responses \cite{gatys2015neural}, represented by gram matrices:
\begin{equation}
G_{ij}^{l} = \sum_{k} F_{ik}^{l} F_{jk}^{l} \label{eq:gram_matrix}
\end{equation}

where \(G_{ij}^{l}\) is the Gram matrix element for filters \(i\) and \(j\) in layer \(l\). The term \(F_{ik}^{l}\) denotes the output of the \(i\)-th filter at position \(k\) in layer \(l\), while \(F_{jk}^{l}\) represents the output of the \(j\)-th filter at the same position and layer. To measure the difference between the content and style images and the generated output image, we use a loss function based on the mean squared error (MSE). For the content loss, we directly measure the difference between the original content image and the target image. It calculates the error rate between the feature maps of the input content image and the output feature maps. The formula for content loss is shown in Equation \eqref{eq:content_loss}.

\begin{equation}
L_{\text{content}}(p, x, l) = \frac{1}{2} \sum_{i,j}(F_{ij}^{l} - P_{ij}^{l})^2 \label{eq:content_loss}
\end{equation}

where $L_{\text{content}}(p, x, l)$ is the content loss between the original image $p$ and the output image $x$ at layer $l$, $F_{ij}^{l}$ is the activation of the $i$-th filter at position $j$ in layer $l$ for $x$, and $P_{ij}^{l}$ is the activation of the $i$-th filter at position $j$ in layer $l$ for $p$.

The style loss is defined by comparing the style representation of the generated image with that of the reference style image through the Gram matrix, which captures the correlations between feature maps across layers. For a given layer $l$, the contribution to the style loss is expressed as

\begin{equation}
E_l = \frac{1}{4 N_l^2 M_l^2} \sum_{i,j} (G_{ij}^{l} - A_{ij}^{l})^2,
\end{equation}

where $N_l$ is the number of filters in layer $l$, $M_l$ is the size of the feature map, $G_{ij}^{l}$ is the Gram matrix computed from the original style image $a$, and $A_{ij}^{l}$ is the Gram matrix obtained from the generated image $x$. The overall style loss is then obtained by summing the weighted contributions across selected layers:

\begin{equation}
L_{\text{style}}(a, x) = \sum_{l=0}^{L} w_l E_l,
\end{equation}

where $w_l$ is the weighting factor for layer $l$. Finally, the total loss combines the content and style losses, balancing the structural fidelity of the original image with the stylistic patterns of the target:

\begin{equation}
L_{\text{total}}(p, a, x) = \alpha L_{\text{content}}(p, x) + \beta L_{\text{style}}(a, x),
\end{equation}

where $\alpha$ and $\beta$ are hyperparameters that control the relative importance of the content and style components, respectively.

Subsequently, the gradients of this mean squared error loss are computed during backpropagation to update the network parameters. The total loss is typically formed by combining the content loss and style loss, each of which is calculated using MSE, which guides the optimization to generate an image that simultaneously preserves the content and captures the style of the target images.

\subsection{Pretrained CNN Models}
Pretrained CNN architectures were used as feature extractors to evaluate their effectiveness in preserving content structure and transferring stylistic characteristics in batik images. The evaluated models included ResNet50, ResNet101, VGG-16, VGG-19, and Inception-v3, and ResNet architectures. ResNet was chosen for its well-established balance between high-quality feature representation and computational efficiency, while VGG and Inception-v3 were included to capture complementary texture and multi-scale features. Our experimental design comprised 150 main experiments with 30 samples per model, supplemented by 49 ablation studies investigating hyperparameter sensitivity. 
We assessed performance through both quantitative metrics (SSIM, PSNR, LPIPS, MSE, training time) and qualitative visual analysis across multiple epoch progressions.

\subsubsection{VGG}
\begin{figure}[h!]
    \centering
    \includegraphics[width=0.9\textwidth]{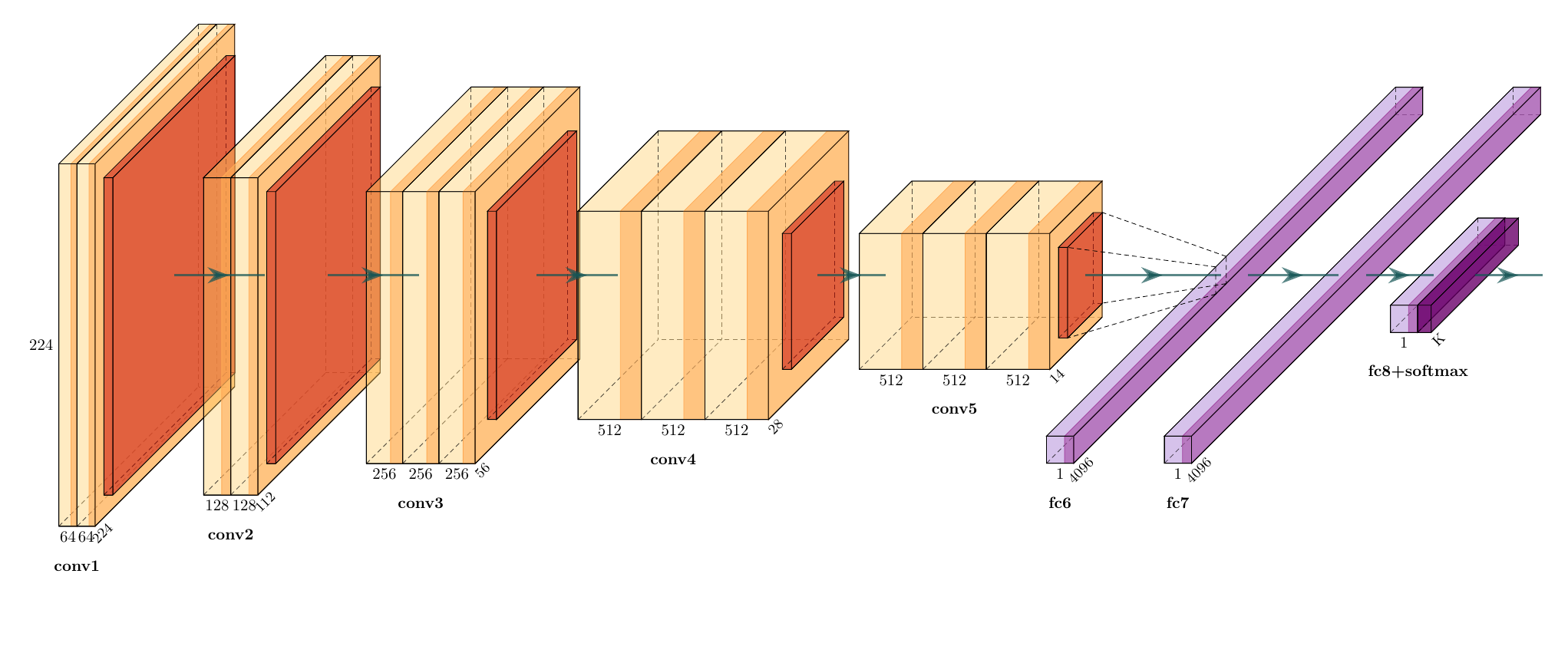}
    \caption{Architecture of VGG-16}
    \label{fig:vgg}
\end{figure}

VGG is a deep convolutional neural network architecture introduced by the Visual Geometry Group at Oxford University~\cite{simonyan2015verydeepconvolutional}. The model was designed to investigate the impact of network depth on large-scale image recognition tasks. Its primary contribution lies in demonstrating that increasing the depth of the network to 16–19 layers, while using very small $3 \times 3$ convolutional filters, yields significant performance improvements over previous state-of-the-art models. VGG-19 achieved outstanding results in the ImageNet Challenge 2014, securing first and second places in the localization and classification tracks, respectively. The architecture of VGG-19 can be seen in Figure~\ref{fig:vgg}.

The use of small convolutional kernels in deep stacked layers allows VGG-19 to capture fine-grained image representations while maintaining manageable computational complexity. Due to its proven capability in extracting hierarchical visual features, VGG-19 was adopted by Gatys et al.~\cite{gatys2015neural} as the backbone for the original neural style transfer framework. In this study, VGG-19 is used as the baseline model, serving as a reference point to compare the performance of more advanced CNN architectures such as ResNet-50 and Inception V3.

\subsubsection{ResNet-101}
ResNet-101 is a convolutional neural network architecture based on the residual learning framework proposed by He et al.~\cite{7780459}. 
The model was introduced to address the optimization difficulties in very deep networks, where increasing depth often leads to performance degradation. 
The core idea of ResNet is the use of \textit{residual connections}, which allow the network to learn residual mappings with respect to the inputs instead of directly fitting the desired functions. 

\begin{figure}[h!]
    \centering
    \includegraphics[width=0.9\textwidth]{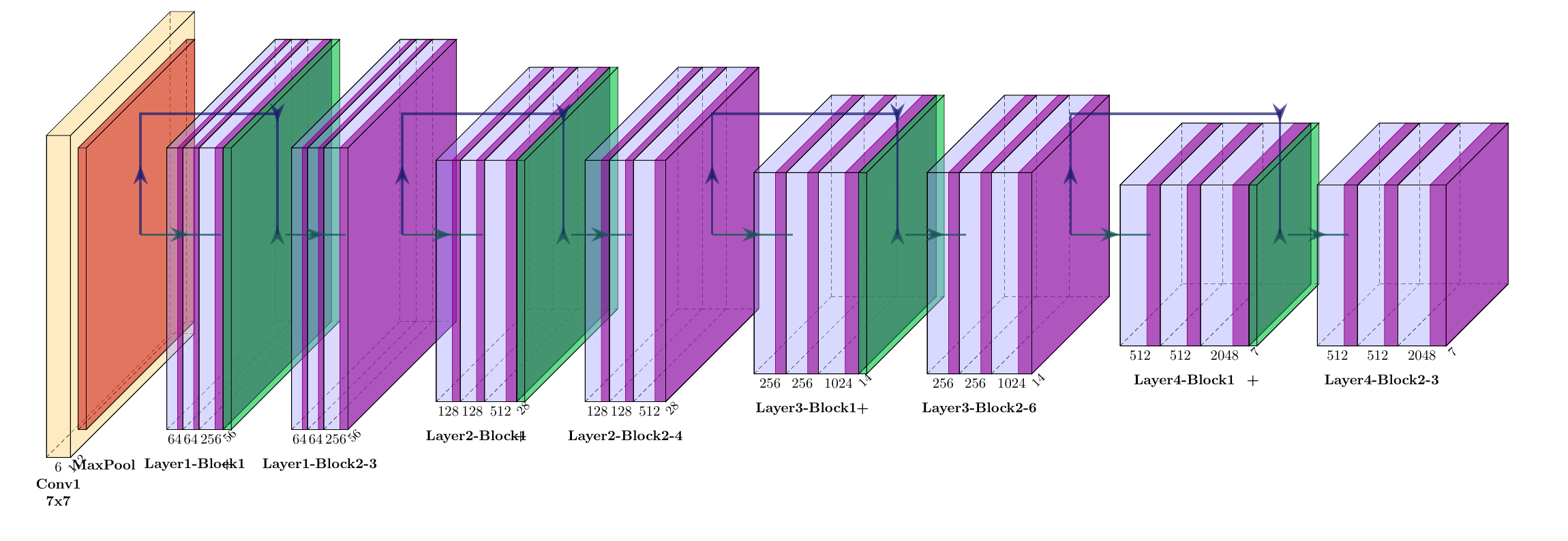}
    \caption{Architecture of ResNet}
    \label{fig:resnet}
\end{figure}

This simple but powerful reformulation enables the successful training of extremely deep architectures, with ResNet achieving state-of-the-art results on ImageNet, 
including a 3.57\% top-5 error rate and winning first place in the ILSVRC 2015 classification task. We used the standard ResNet implementations from torchvision.models, accessing intermediate features through layer-specific hooks. The residual connections, while beneficial for classification tasks, introduce architectural complexity for style transfer, as features propagate through both convolutional paths and skip connections.
\subsubsection{Inception V3}
Inception~V3, developed by Szegedy \textit{et al.}~\cite{inceptionmain}, employs a multi-scale architecture in which parallel convolutional pathways process inputs at different receptive field sizes (1$\times$1, 3$\times$3, and 5$\times$5 filters) within each Inception module. 
This design enables efficient extraction of features at multiple spatial scales simultaneously. 
In this study, Inception~V3 was implemented using the \texttt{torchvision} pre-trained model, with auxiliary classifiers disabled during the style transfer process to reduce computational overhead. 

\begin{figure}[h!]
    \centering
    \includegraphics[width=0.9\textwidth]{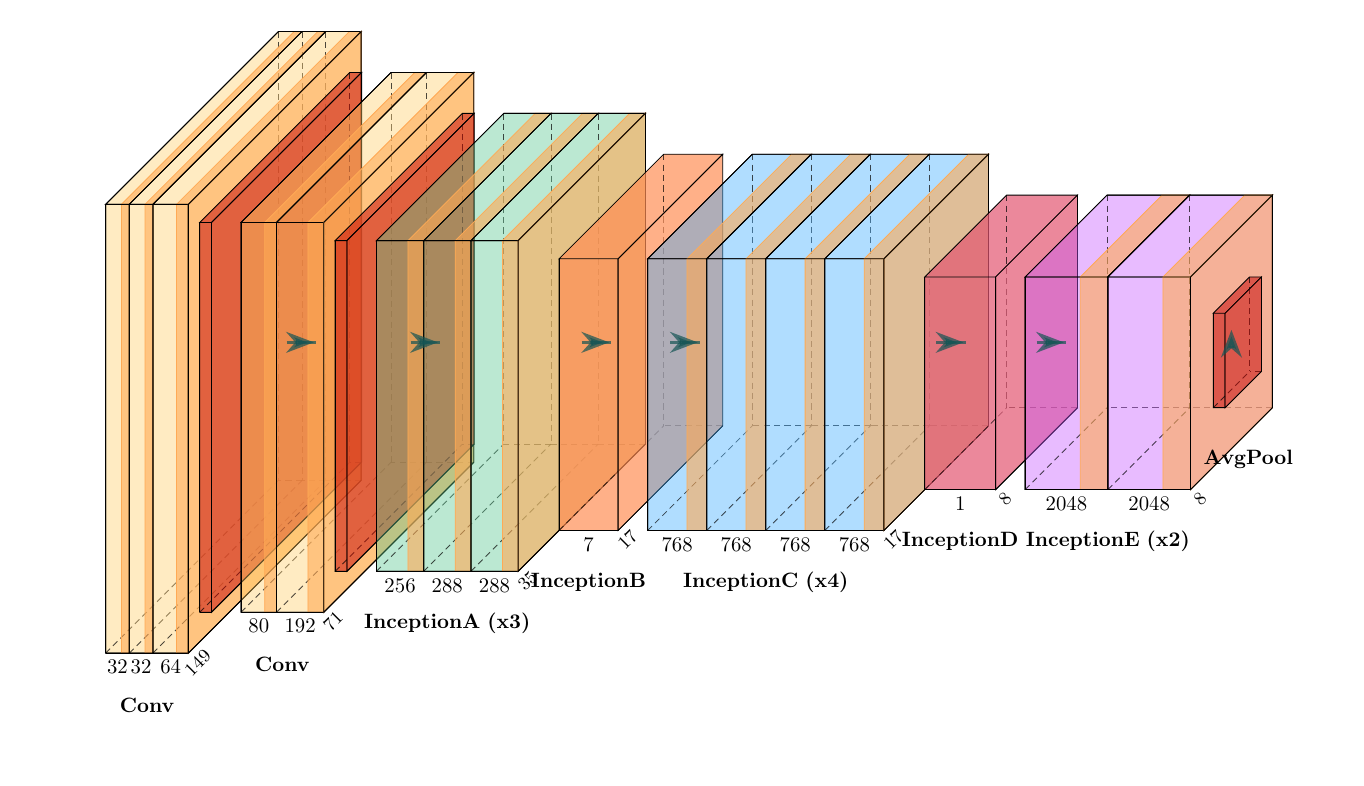}
    \caption{Architecture of Inception}
    \label{fig:inception}
\end{figure}

The presence of parallel pathways in each Inception block creates a complex feature space, which can either facilitate or hinder style transfer performance depending on how features from different scales interact during the optimization process.

\subsection{Materials}

\subsection{Evaluation Metrics}
We assessed style transfer quality using four complementary quantitative metrics, each capturing a distinct aspect of output quality, in addition to qualitative visual analysis.

\textbf{Structural Similarity Index (SSIM).}
SSIM measures perceptual similarity between two images based on luminance, contrast, and structural information:
\begin{equation}
\mathrm{SSIM}(x, y) = \frac{(2 \mu_x \mu_y + C_1)(2 \sigma_{xy} + C_2)}{(\mu_x^2 + \mu_y^2 + C_1)(\sigma_x^2 + \sigma_y^2 + C_2)},
\end{equation}
where $\mu_x$ and $\mu_y$ denote the mean intensities of images $x$ and $y$, $\sigma_x^2$ and $\sigma_y^2$ their variances, $\sigma_{xy}$ their covariance, and $C_1$ and $C_2$ are small constants introduced for numerical stability. 
In this study, SSIM was computed between the content image and the generated output using a sliding window approach with an 11$\times$11 Gaussian kernel. 
SSIM values range from $-1$ to $1$, with higher values indicating stronger structural preservation, typically falling within the $0.2$--$0.5$ range for style transfer tasks.

\textbf{Peak Signal-to-Noise Ratio (PSNR).}
PSNR quantifies the ratio between the maximum possible signal power and the power of corrupting noise, derived from the mean squared error between the content and output images:
\begin{equation}
\mathrm{PSNR} = 10 \log_{10} \left( \frac{MAX^2}{\mathrm{MSE}} \right),
\end{equation}
where $MAX$ denotes the maximum possible pixel value, set to $1.0$ for normalized images. 
Higher PSNR values indicate closer pixel-level similarity to the content image, alongside the perceptual emphasis of SSIM.

\textbf{Learned Perceptual Image Patch Similarity (LPIPS).} LPIPS measures perceptual distance using deep feature representations extracted from a pre-trained network~\cite{zhang2018perceptual}. For consistency, a VGG-based backbone was employed. Unlike SSIM and PSNR, which operate directly on pixel intensities, LPIPS evaluates similarity in a learned feature space that aligns more closely with human visual perception. 

Importantly, LPIPS in this study is computed between the generated output and the original content image. Therefore, lower LPIPS values indicate stronger content preservation (i.e., the output retains more structural similarity to the content), while higher LPIPS values suggest greater stylistic transformation. This interpretation is critical: LPIPS does not directly measure ``quality'' in the artistic sense, but rather quantifies the degree of perceptual deviation from the content baseline. Typical style transfer outputs range from 0.3 to 0.7, with values closer to 0.3, that indicate conservative stylization and values near 0.7 indicating more aggressive content modification.

\textbf{Mean Squared Error (MSE).}
MSE provides a simple pixel-wise difference measure, defined as:
\begin{equation}
\mathrm{MSE} = \frac{1}{N} \sum_{i=1}^{N} (x_i - y_i)^2,
\end{equation}
where $N$ denotes the total number of pixels, and $x_i$ and $y_i$ represent corresponding pixel intensities in the content and output images. 
Although less perceptually meaningful, MSE serves as a baseline metric and maintains a direct mathematical relationship with PSNR.

\textbf{Training Time.}
Training time was recorded as wall-clock time in seconds, measured from optimization initialization to final convergence or the maximum epoch limit. 
This metric captures computational efficiency, which is critical for practical deployment scenarios.

All metrics were computed using Python implementations: SSIM and PSNR from \texttt{scikit-image}, LPIPS from the \texttt{lpips} package, and MSE from \texttt{NumPy}. 
Metrics were evaluated on the final output image at epoch 5000, with additional evaluations at epochs 100 and 2500 for convergence analysis.

\subsection{Experimental Setup} 
All experiments were conducted on a single NVIDIA GeForce GTX~1650 GPU with 4~GB VRAM. This configuration was deliberately chosen to reflect a constrained yet commonly available computational setting, rather than an idealized high-end research environment. By fixing both memory and compute resources across all experiments, we ensure fair and reproducible architectural comparisons, which enables differences in convergence behavior, runtime efficiency, and training stability to be evaluated under realistic resource limitations.

Our baseline hyperparameter configuration was established through preliminary experiments and represents a balanced setting for batik style transfer. As detailed in Table~\ref{tab:baseline_config}, the large style weight relative to content weight (ratio $1:10^8$) reflects the need to strongly prioritize style transfer given the magnitude difference between content and style loss values. This ratio was empirically determined to produce visually balanced outputs where batik patterns are clearly visible while content structure remains recognizable.

\begin{table}[H]
    \centering
    \caption{Baseline hyperparameter configuration used for the primary style transfer experiments.}
    \label{tab:baseline_config}
    
    \small 
    
    \begin{tabular}{lcl}
        \toprule
        \textbf{Parameter} & \textbf{Value} & \textbf{Description} \\
        \midrule
        Content Weight ($\alpha$) & $1.0$ & Standard content reference \\
        Style Weight ($\beta$)    & $1 \times 10^8$ & Strong style priority \\
        Content Layer             & Layer 2 & Mid-level features \\
        Style Layer               & Layer 8 & Higher-level textures \\
        Learning Rate             & 0.05 & Balanced speed/stability \\
        Max Epochs                & 5000 & Converged training \\
        Pooling                   & Original & Pre-trained default \\
        \bottomrule
    \end{tabular}
\end{table}

For the ablation study, we systematically varied these parameters while holding others constant to analyze sensitivity. We designed three distinct experimental sets: Content/Style Weight Ratios, Layer Selection, and Learning Rate variations. The specific configurations for these ablation sets are summarized in Table~\ref{tab:ablation_setup}. Each experiment used a fixed subset of 3 representative image pairs to enable rapid iteration while maintaining statistical validity.

\begin{table}[H]
    \centering
    \caption{Design of Ablation Study Experiments showing variations in Weights, Layers, and Learning Rates.}
    \label{tab:ablation_setup}
    \resizebox{\columnwidth}{!}{%
    \begin{tabular}{lll}
        \toprule
        \textbf{Experiment Set} &\textbf{ Variant Name} & \textbf{Configuration Details} \\
        \midrule
        \multicolumn{3}{l}{\textbf{Set 1: Content/Style Weight Ratios (9 exp.)}} \\
         & \textit{Baseline} & $\alpha=1, \ \beta=1 \times 10^8$ \\
         & Variant A & $\alpha=1, \ \beta=1 \times 10^7$ (Reduced style) \\
         & Variant B & $\alpha=1, \ \beta=1 \times 10^9$ (Increased style) \\
         & Variant C & $\alpha=10, \ \beta=1 \times 10^8$ (Increased content) \\
        \midrule
        \multicolumn{3}{l}{\textbf{Set 2: Layer Selection Combinations (9 exp.)}} \\
         & \textit{Baseline} & Content: L2, Style: L8 \\
         & Shallow Layers & Content: L1, Style: L6 \\
         & Deep Layers & Content: L3, Style: L10 \\
         & Multi-layer & Content: L2, Style: [6, 8, 10] \\
        \midrule
        \multicolumn{3}{l}{\textbf{Set 3: Learning Rate Variations (4 exp.)}} \\
         & \textit{Baseline} & LR = 0.05 \\
         & Conservative & LR = 0.01 (Stable, slow) \\
         & Aggressive & LR = 0.1 (Fast, potentially unstable) \\
         & Very Aggressive & LR = 0.2 (Limit testing) \\
        \bottomrule
    \end{tabular}%
    }
\end{table}
\section{Results}
In this section, we present a comprehensive evaluation of five convolutional neural network architectures such as VGG16, VGG19, Inception V3, ResNet50, and ResNet101 for Neural Style Transfer applied to Indonesian batik motives. Our experimental design comprised 245 experiments across 49 image pairs per model, with comprehensive assessment through quantitative metrics (SSIM, PSNR, LPIPS, training time) and qualitative visual analysis.

\subsection{Quantitative Performance Analysis}
We assessed each architecture's ability to balance content preservation with style transfer effectiveness through multiple quantitative metrics. Table \ref{tab:quantitative_results} presents the comprehensive results across all evaluated models.

\begin{table}[H]
\centering
\caption{Quantitative performance comparison across architectures}
\label{tab:quantitative_results}
\resizebox{\linewidth}{!}{%
\begin{tabular}{lcccc}
\toprule
\textbf{Architecture} &
\textbf{SSIM$\uparrow$} &
\textbf{PSNR (dB)$\uparrow$} &
\textbf{LPIPS} &
\textbf{Training Time (s) $\downarrow$} \\
\midrule
\textbf{ResNet50} & \textbf{0.297$\pm$0.110} & \textbf{10.81$\pm$2.69} & 0.527$\pm$0.071 & 25.4$\pm$3.8 \\
Inception V3 & 0.293$\pm$0.104 & 9.89$\pm$2.08 & 0.654$\pm$0.054 & 71.0$\pm$10.4 \\
VGG19 & 0.288$\pm$0.108 & 9.36$\pm$2.03 & 0.635$\pm$0.059 & 157.7$\pm$61.4 \\
VGG16 & 0.281$\pm$0.106 & 9.32$\pm$2.02 & 0.640$\pm$0.052 & 129.5$\pm$25.2 \\
ResNet101 & 0.274$\pm$0.105 & 10.69$\pm$2.60 & 0.540$\pm$0.075 & \textbf{23.3$\pm$3.8} \\
\bottomrule
\end{tabular}%
}
\end{table}

Our quantitative evaluation indicates that differences in visual quality metrics across architectures are relatively modest. As summarized in Table~\ref{tab:quantitative_results}, SSIM values cluster within a narrow range (0.27--0.30) for all evaluated models, with ResNet50 achieving the highest mean SSIM (0.297~$\pm$~0.110), followed closely by Inception~V3 (0.293~$\pm$~0.104) and VGG19 (0.288~$\pm$~0.108). This findings confirms that these differences are not significant ($p = 0.83$), indicating comparable levels of structural content preservation across architectures. Importantly, higher SSIM values should be interpreted as stronger preservation of content structure rather than superior stylization quality. In this context, the slightly higher SSIM observed for ResNet-based models reflects milder stylistic deviation from the content image, whereas VGG-based architectures introduce more pronounced textural transformations, which led in lower structural similarity.

While visual quality remains statistically comparable, architectural differences manifest strongly in computational efficiency. A key finding of this study is that ResNet-based architectures are able to match the perceptual and structural outcomes of VGG-based NST while operating at substantially lower computational cost. As shown in Table~\ref{tab:quantitative_results}, ResNet50 and ResNet101 complete optimization approximately $5$--$6\times$ faster than VGG16 and VGG19, and around $3\times$ faster than Inception~V3 under identical experimental conditions. This efficiency advantage is particularly noteworthy given that VGG architectures are widely regarded as the de facto standard for NST quality \cite{9918891, pihlgren2024systematicperformanceanalysisdeep, 10578183}.

Perceptual similarity metrics further reinforce this observation. As detailed in Table \ref{tab:quantitative_results}, ResNet50 achieves the lowest LPIPS score (0.527$\pm$0.071), which shows \textit{stronger content preservation} compared to VGG and Inception variants, which exhibit higher LPIPS values in the 0.635--0.654 range. VGG models, conversely, demonstrate higher LPIPS values (0.635--0.654), signifying greater perceptual deviation from the content image. This corresponds to VGG`s tendency to produce denser, more painterly texturization that substantially modifies the original content structure in favor of expressive stylistic effects. Inception V3 occupies an intermediate position (LPIPS $\approx$ 0.64), which indicates a balance between content preservation and stylistic transformation, though with increased visual noise artifacts.

PSNR follows a consistent trend, with ResNet models slightly surpassing 10~dB on average, compared to approximately 9.3--9.9~dB for VGG and Inception architectures. The higher PSNR values for ResNet further corroborate its conservative approach to content modification, maintaining closer pixel-level correspondence with the original content.

Taken together, these results demonstrate that the dramatic improvement in computational efficiency offered by ResNet backbones (5--6$\times$ faster training, 16$\times$ fewer FLOPs) is achieved through a different stylistic strategy rather than at the expense of output quality. ResNet prioritizes structural preservation with efficient stylization, while VGG favors expressive transformation with higher computational cost. The choice between architectures should therefore be guided by application requirements: ResNet for structure-preserving efficiency, VGG for maximal artistic expressiveness.

\begin{figure}[H]
    \centering
    \includegraphics[width=\linewidth]{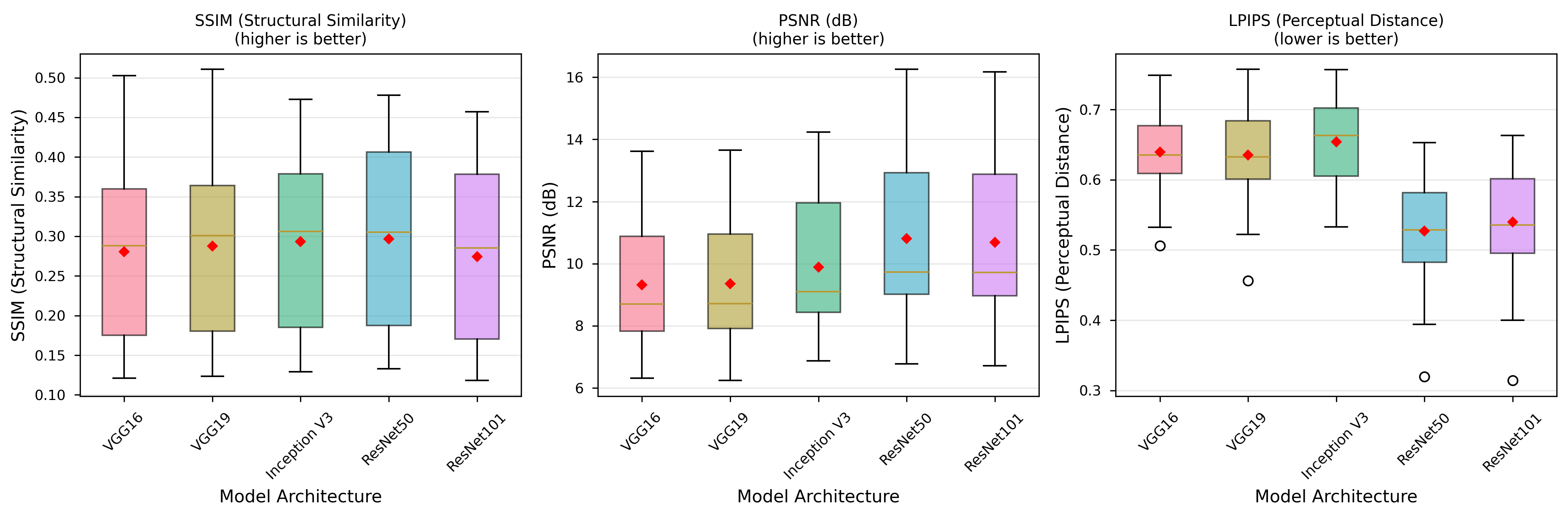}
    \caption{Box plot distribution of quantitative performance metrics (SSIM, PSNR, and LPIPS) across different CNN architectures.}
    \label{fig:metrics_boxplots}
\end{figure}

Variance analysis across image pairs shows broadly similar dispersion patterns among architectures, as illustrated in the boxplots in Fig.~\ref{fig:metrics_boxplots}. The distributions for SSIM, PSNR, and LPIPS exhibit substantial overlap, further suggesting that architectural choice does not induce systematic instability or quality degradation. While ResNet models display slightly wider SSIM interquartile ranges (IQR $\approx 0.18$ vs. $0.15$ for VGG), this likely reflects greater sensitivity to content--style interactions across diverse motive geometries rather than performance instability. Crucially, this variability remains within the same overall performance envelope as the more computationally expensive VGG-based approaches.

When considered collectively, the quantitative metrics indicate that architectural choice does not lead to a single dominant model across all evaluation criteria. While VGG-based architectures are historically favored for their strong stylistic emphasis, our data suggests this advantage is accompanied by a prohibitive computational cost ($>5\times$ slower). In contrast, ResNet50 preserves comparable levels of structural and perceptual fidelity while requiring significantly less computation. This establishes ResNet50 as yielding a superior efficiency--quality trade-off, making it the optimal candidate for resource-aware industrial settings where throughput is as critical as aesthetic fidelity.

\subsection{Statistical Significance Analysis}
To assess whether the observed performance differences represent meaningful architectural advantages, we conducted statistical analyses. One-way ANOVA on SSIM yielded $F(4, 240) = 0.37$, $p = 0.83$,  which shows no statistically significant differences among the five architectures at the $\alpha = 0.05$ level. The high p-value ($p = 0.83$) suggests an 83\% probability that the observed differences could arise from random variation alone, rather than genuine architectural effects, as shown in Table~\ref{tab:anova_ssim}.

\begin{table}[H]
\centering
\caption{One-way ANOVA results for SSIM across architectures}
\label{tab:anova_ssim}
\resizebox{\linewidth}{!}{%
\begin{tabular}{lccccc}
\toprule
\textbf{Source} & \textbf{df} & \textbf{Sum of Squares} & \textbf{Mean Square} & \textbf{F-statistic} & \textbf{p-value} \\
\midrule
Between Groups & 4   & 0.0175 & 0.0044 & 0.37 & 0.832 \\
Within Groups  & 240 & 2.8320 & 0.0118 & --   & --    \\
Total          & 244 & 2.8495 & --     & --   & --    \\
\bottomrule
\end{tabular}%
}
\end{table}

Following the non-significant ANOVA result, the pairwise analyses, as shown in Table~\ref{tab:pairwise_ssim}, revealed no meaningful differences between the architectures. All comparisons produced $p$-values above $0.05$, and the largest observed difference between ResNet50 and ResNet101 ($\Delta = 0.023$) remained statistically non-significant ($p = 0.30$). The small effect size ($\eta^2 = 0.006$), calculated as $\mathrm{SS}_{\text{between}} / \mathrm{SS}_{\text{total}} = 0.0175 / 2.849$, indicates that architectural choice explains less than $1\%$ of the variance in SSIM performance, which suggests that other factors, such as hyperparameters, image characteristics, and training dynamics, have a greater influence on the final output quality.

\begin{table}[H]
\centering
\caption{Pairwise t-test results for SSIM comparisons (selected pairs)}
\label{tab:pairwise_ssim}
\resizebox{\linewidth}{!}{%
\begin{tabular}{llccccc}
\toprule
\textbf{Model 1} & \textbf{Model 2} & \textbf{Mean Diff} & \textbf{t-statistic} & \textbf{p-value} & \textbf{Cohen's $d$} & \textbf{Significant} \\
\midrule
ResNet50 & ResNet101       & +0.023 & 1.04  & 0.300 & 0.21  & No \\
ResNet50 & VGG16           & +0.016 & 0.74  & 0.460 & 0.15  & No \\
ResNet50 & VGG19           & +0.009 & 0.42  & 0.674 & 0.09  & No \\
ResNet50 & Inception V3    & +0.004 & 0.16  & 0.872 & 0.03  & No \\
VGG19    & VGG16           & -0.007 & -0.32 & 0.749 & -0.06 & No \\
ResNet101& Inception V3    & +0.019 & 0.91  & 0.366 & 0.18  & No \\
\bottomrule
\end{tabular}%
}
\end{table}

The absence of statistical significance indicates that architectural choice does not lead to measurable differences in structural similarity as quantified by SSIM. However, this result should not be interpreted as evidence of identical perceptual quality or stylistic intensity across architectures. Rather, the SSIM equivalence reflects a comparable degree of structural preservation with respect to the content image. In this context, higher SSIM values obtained by ResNet-based models indicate stronger retention of the original geometric layout, rather than superior visual richness.

This interpretation is consistent with the qualitative observations, where VGG architectures produce denser and more pronounced painterly textures, while ResNet backbones exhibit comparatively milder stylization with greater structural stability. Consequently, ResNet models should not be interpreted as achieving equivalent stylistic quality, but as offering a distinct trade-off between reduced stylistic intensity and improved structural preservation together with computational efficiency. While this behavior may appear under-stylized from an artistic perspective, it is advantageous for heritage-oriented applications where fidelity to original line structures is prioritized.

\begin{figure}[H]
    \centering
    \includegraphics[width=\linewidth]{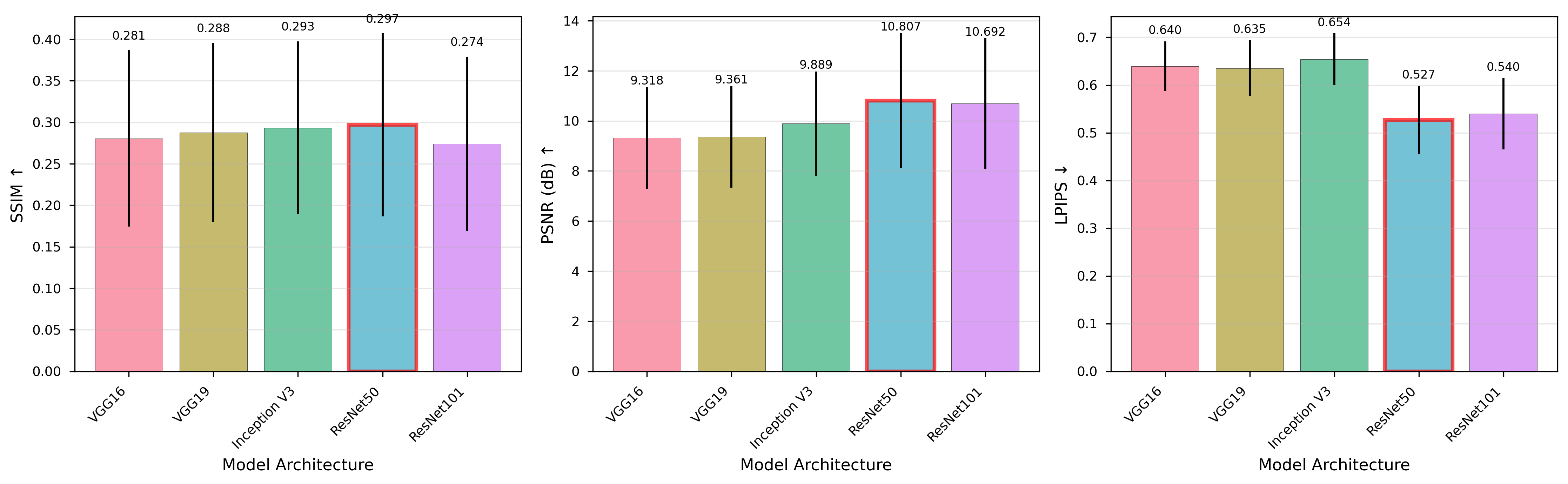}
    \caption{Mean Comparison Plot of quantitative performance metrics (SSIM, PSNR, and LPIPS) across different CNN architectures.}
    \label{fig:metrics_barplot}
\end{figure}

From an applied standpoint, these findings suggest that architectural selection in NST should be guided by the intended balance between stylistic expressiveness and structural fidelity, rather than by SSIM alone. The consistent metric rankings, where ResNet50 attains the highest SSIM and lowest LPIPS, combined with substantial reductions in training time, provide actionable guidance for scenarios in which efficiency and structure preservation are more critical than maximal stylization strength. A detailed comparison of metric distributions is shown in Figure~\ref{fig:metrics_barplot}.

Mean performance metrics with standard deviation error bars indicate that ResNet50 achieves marginally higher SSIM (0.297) and lower LPIPS (0.527), however, these differences remain small relative to the corresponding error bars.

\begin{figure}[H]
    \centering
    \includegraphics[width=\linewidth]{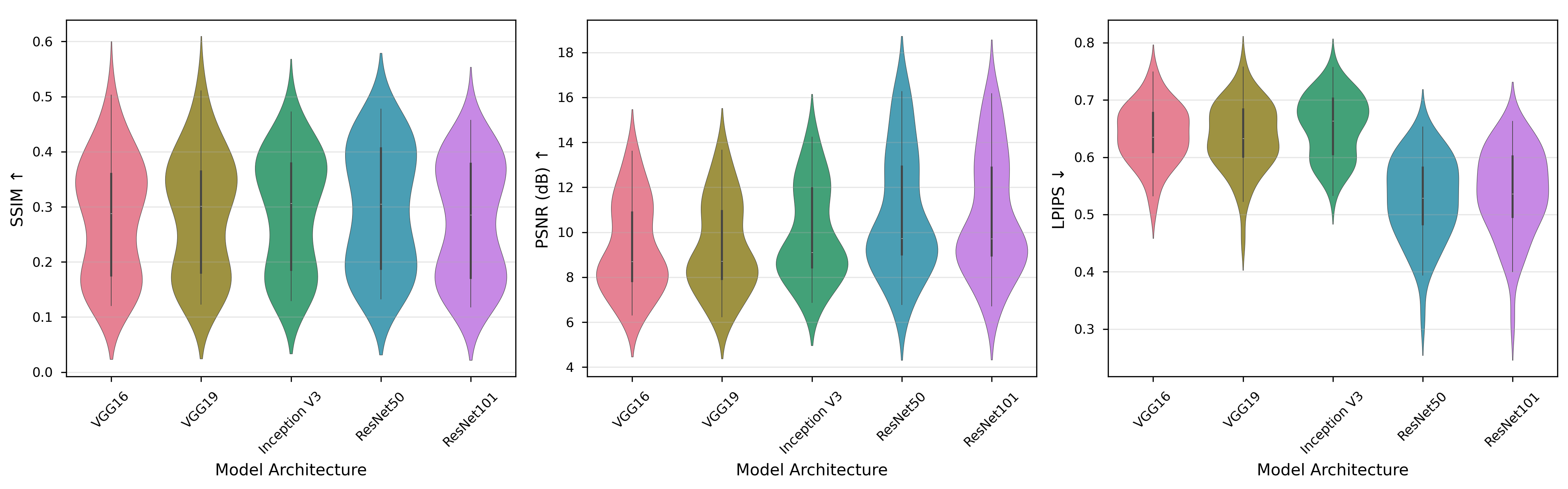}
    \caption{Distribution of quantitative performance metrics (SSIM, PSNR, and LPIPS across different CNN architectures.}
    \label{fig:metrics_violinplot}
\end{figure}

As shown in Fig.~\ref{fig:metrics_violinplot}, the distributions of SSIM, PSNR, and LPIPS largely overlap across architectures, which indicates that quantitative differences in visual quality are relatively modest. The largely symmetric and unimodal distributions indicate stable performance across image pairs, with no evidence of architecture-specific failure modes. However, this statistical similarity should not be interpreted as equivalence in stylistic behavior.

\subsection{Visual Quality Assessment}
\begin{figure}[H]
    \centering
    \includegraphics[width=\linewidth]{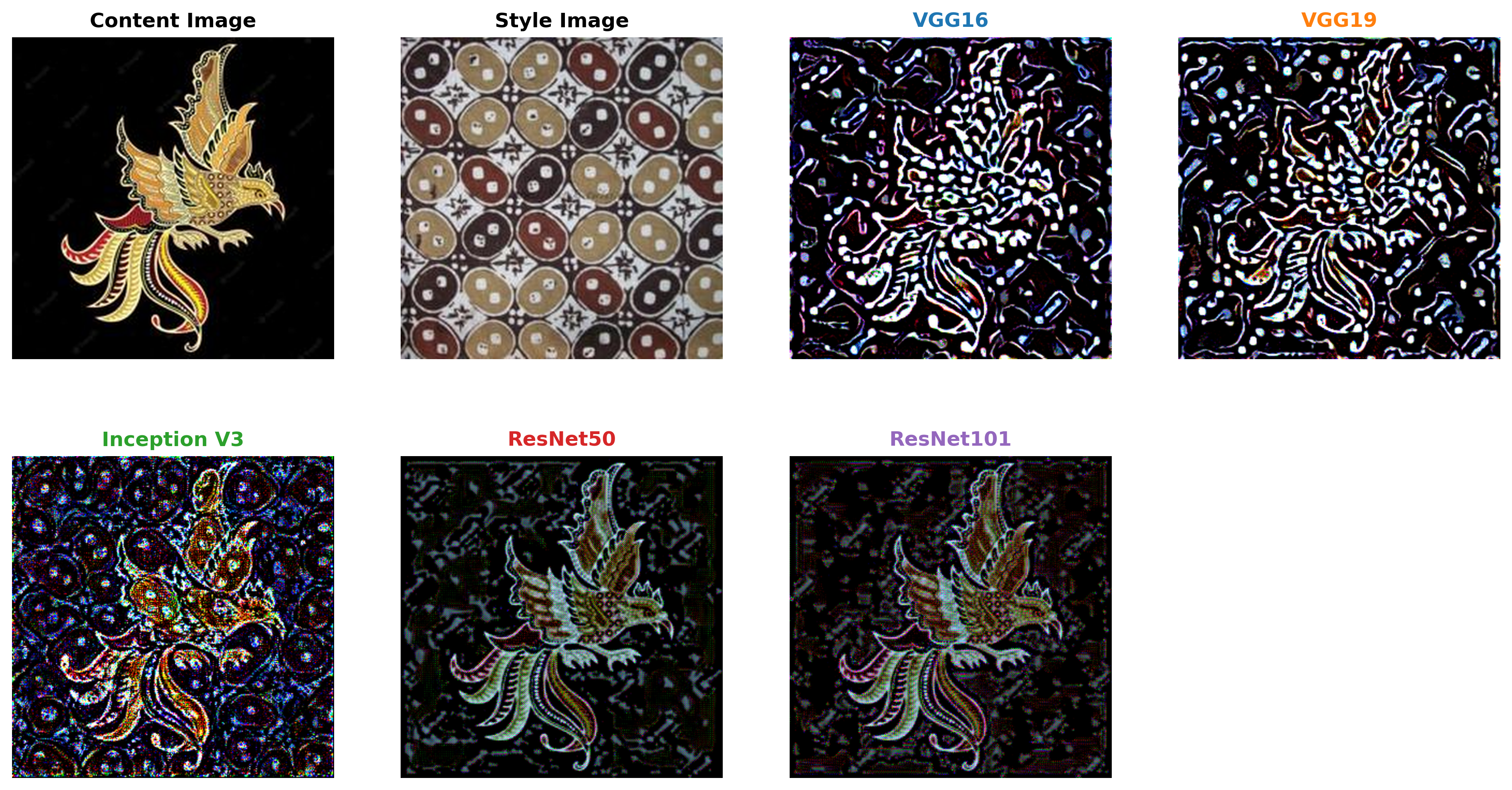}
    \caption{Combination of content image and style image based on each architecture results.}
    \label{fig:visual_comparison_pair28}
\end{figure}

In addition to the quantitative results, visual inspection reveals clear differences in how backbone architectures handle style and structure. These differences can be observed in Figure~\ref{fig:visual_comparison_pair28}, which compares VGG16, VGG19, Inception~V3, ResNet50, and ResNet101 under identical content and style inputs.

The chosen example features a clearly defined foreground combined with a highly dynamic batik-style pattern that makes differences in stylization. When VGG-based feature extractors are applied, stylistic elements quickly dominate the image. Both VGG16 and VGG19 generate dense and saturated batik textures that extend across most regions. Although the overall content outline remains visible, fine structural details are partially obscured by the prevalence of high-frequency patterns. Compared to VGG16, VGG19 shows slightly improved texture coherence; however, the images produced in both cases remain strongly driven by stylistic emphasis.

Inception~V3 produces visually rich outputs, yet the integration of stylistic patterns appears less consistent. Texture distribution is uneven, leading to a fragmented visual impression and noticeable noise that reduces the legibility of the underlying structure.

By contrast, ResNet-based architectures adopt a more conservative approach to style transfer. ResNet50 and ResNet101 preserve object boundaries and major contours more effectively, keeping the content structure visually salient. The applied batik style is subtler and more controlled, with ResNet101 exhibiting marginally smoother tonal transitions and a slightly darker overall appearance than ResNet50.

\subsection{Ablation Study on Hyperparameter Sensitivity}

To complement our architectural comparison, we conducted ablation studies to examine the sensitivity of neural style transfer (NST) quality to key hyperparameters, namely content--style weight ratios, feature layer selection, and learning rate. All experiments in this section employ VGG19 as a fixed feature extractor and use a subset of three representative image pairs (Pairs 1, 12, and 20) to enable systematic and controlled comparisons across parameter variations. The results of the ablation study are summarized in Table~\ref{tab:ablation}.

\begin{table}[H]
\centering
\caption{Ablation study results showing mean SSIM, PSNR, and LPIPS across hyperparameter variations}
\label{tab:ablation}
\resizebox{\linewidth}{!}{%
\begin{tabular}{llccc}
\toprule
\textbf{Experiment Set} & \textbf{Configuration} & \textbf{SSIM} & \textbf{PSNR (dB)} & \textbf{LPIPS} \\
\midrule
\multirow{2}{*}{Weight Ratios}
& Baseline ($\alpha=1$, $\beta=1 \times 10^{8}$) & 0.363 & 11.05 & 0.615 \\
& Variant A ($\alpha=1$, $\beta=1 \times 10^{7}$) & 0.427 & 11.46 & 0.575 \\
\midrule
\multirow{4}{*}{Layer Selection}
& Baseline (L2, L8) & 0.363 & 11.05 & 0.615 \\
& Shallow (L1, L6) & 0.375 & 11.00 & 0.622 \\
& Deep (L3, L10) & 0.365 & 11.43 & 0.605 \\
& Multi-layer (L2, {[6, 8, 10]}) & 0.326 & 10.56 & 0.639 \\
\midrule
\multirow{3}{*}{Learning Rate}
& Baseline (LR = 0.05) & 0.363 & 11.05 & 0.615 \\
& Conservative (LR = 0.01) & 0.496 & 13.48 & 0.553 \\
& Aggressive (LR = 0.1) & 0.287 & 10.49 & 0.641 \\
\bottomrule
\end{tabular}%
}
\end{table}

\textbf{Content--Style Weight Ratios}
We evaluated three variations of the content--style balance relative to our baseline configuration ($\alpha = 1$, $\beta = 1 \times 10^{8}$). Reducing the style weight to $\beta = 1 \times 10^{7}$ (Variant A) improved the mean SSIM from 0.363 to 0.427, which suggests that the baseline setting may apply excessive stylization for certain batik patterns. Meanwhile, increasing the style weight to $\beta = 1 \times 10^{9}$ (Variant B) or increasing the content weight to $\alpha = 10$ (Variant C) resulted in negligible quality improvements, which indicate diminishing returns beyond the baseline ratio. These results imply that a moderate reduction in style weight can enhance content preservation while maintaining sufficient stylistic pattern transfer.

\textbf{Feature Layer Selection}
The depth of selected feature layers strongly influences the abstraction level of the transferred style. Using shallow layers (content: L1, style: L6) yielded a mean SSIM of 0.375, capturing fine-grained textures but potentially over-emphasizing low-level patterns. Conversely, deeper layers (content: L3, style: L10) produced a mean SSIM of 0.365, with an emphasis on semantic content preservation over detailed texture reproduction. Employing multi-layer style extraction (L6, L8, L10) resulted in a lower SSIM of 0.326, suggesting that combining multiple abstraction levels may introduce conflicting stylistic signals for repetitive batik patterns. In this setting, the baseline mid-level configuration (content: L2, style: L8) represents a reasonable compromise between texture fidelity and semantic coherence.

\textbf{Learning Rate Impact}
Among the evaluated hyperparameters, the learning rate emerged as the most influential factor affecting NST quality. A conservative learning rate (LR = 0.01) achieved the highest mean SSIM of 0.496, substantially outperforming the baseline configuration (LR = 0.05, SSIM = 0.363). This improvement indicates that slower and more stable optimization facilitates better convergence toward high-quality local minima. In contrast, an aggressive learning rate (LR = 0.1) degraded performance to a mean SSIM of 0.287, likely due to optimization instability and overshooting of optimal solutions. Notably, training time remained comparable across learning rates (approximately 137 seconds per run), this suggests that the quality--efficiency trade-off favors conservative learning rates for batik style transfer.

These results indicate that hyperparameter choices substantially influence the visual characteristics of batik neural style transfer (NST) outputs. The observed SSIM variation across configurations (0.287 to 0.496) reflects differing degrees of style transfer intensity. Conservative learning rates (LR = 0.01) result in weaker stylization, with outputs remaining closer to the original content structure (SSIM = 0.496), whereas more aggressive rates (LR = 0.1) enable stronger batik pattern application (SSIM = 0.287). Similarly, reducing the style weight ($\beta = 1 \times 10^{7}$) produces less pronounced stylization (SSIM = 0.427) compared to the baseline configuration (SSIM = 0.363).

These results further suggest that hyperparameter tuning provides practitioners with direct control over the visual appearance of batik style transfer. In contrast to architectural selection, which exhibited a relatively narrow impact on style transfer intensity (SSIM range 0.27--0.30) adjustments to learning rate and content--style weight ratios enable substantial variation in how prominently batik patterns are applied. The choice of configuration therefore depends on user preference and intended aesthetic, ranging from subtle batik integration (higher SSIM, stronger content preservation) to bold pattern transformation (lower SSIM, increased stylization). Neither extreme can be considered objectively superior; rather, each supports distinct artistic intentions and application contexts.

\subsection{Training Efficiency and Computational Cost Analysis}

\subsubsection{Computational Requirements}
To better understand the observed training time differences, we analyzed the computational requirements of each architecture. Table~\ref{tab:computational_cost} presents a comprehensive comparison of model parameters, floating-point operations (FLOPs), GPU memory usage, and inference time.

\begin{table}[H]
\centering
\caption{Computational requirements comparison across architectures}
\label{tab:computational_cost}
\makebox[\linewidth][c]{
\begin{tabular}{lrrrr}
\hline
\textbf{Architecture} & \textbf{Params (M)} & \textbf{FLOPs (G)} & \textbf{VRAM (GB)} & \textbf{Time (ms)} \\
\hline
VGG19 & 20.02 & 10.12 & 0.203 & 9.24 \\
VGG16 & \textbf{14.71} & 10.12 & 0.184 & 9.04 \\
ResNet50 & 23.51 & \textbf{0.63} & \textbf{0.122} & \textbf{1.28} \\
ResNet101 & 42.50 & \textbf{0.63} & 0.193 & 1.33 \\
Inception V3 & 21.79 & 4.08 & 0.123 & 4.64 \\
\hline
\end{tabular}
}
\end{table}

The computational analysis reveals a striking paradox: VGG architectures, despite having fewer parameters (14.71--20.02M) than ResNet models (23.51--42.50M), require substantially more floating-point operations (10.12 GFLOPs vs. 0.63 GFLOPs). This 16$\times$ difference in FLOPs directly explains the 5--6$\times$ training time disparity observed in practice.

ResNet's computational efficiency stems from its residual connection architecture, which enables deeper networks without proportional increases in computational cost. The skip connections allow gradient flow through shorter paths, reducing the number of operations required per forward pass while maintaining representational capacity through increased depth.

VGG's sequential convolutional architecture, conversely, requires processing through every layer without shortcuts, so that in higher computational overhead despite fewer total parameters. The fully connected nature of VGG's feature extraction pipeline necessitates more operations per inference, particularly in deeper layers where feature map dimensions remain large.

Inception V3 demonstrates intermediate computational requirements (4.08 GFLOPs, 4.64ms inference time), which indicates its hybrid design with parallel convolutional paths. While more efficient than VGG, the multi-scale processing introduces overhead compared to ResNet's streamlined residual blocks.

GPU memory requirements remain modest across all architectures (0.12--0.20 GB), showing that VRAM is not a limiting factor for style transfer at 512$\times$512 resolution. The small memory footprint enables batch processing and concurrent execution on consumer-grade hardware.

The strong correlation between FLOPs and observed training time validates our empirical measurements and provides a theoretical foundation for architecture selection. For production deployments requiring high throughput, ResNet's 7$\times$ computational advantage over VGG translates directly to processing capacity, which enables real-time or near-real-time style transfer applications.

\subsubsection{Training Time Distribution}
While quality metrics showed minimal differentiation, training efficiency varied substantially across architectures. We analyzed training time distributions to better understand computational requirements and their practical implications for production workflows.

ResNet architectures completed training in approximately 23--25~s on average with low variance ($\mathrm{std}=3.8$~s). This consistency indicates stable optimization dynamics across diverse image pairs, a valuable property for production environments requiring predictable throughput. The compact distributions with minimal outliers suggest robust convergence regardless of content--style complexity.

VGG models required 130--158~s per image pair, relevant to an approximately 5--6$\times$ increase in training time relative to ResNet. The high variance, particularly for VGG19 ($\mathrm{std}=61.4$~s), with outliers exceeding 250~s, indicates unpredictable convergence behavior. Such variability complicates production scheduling and resource allocation, as certain image pairs require substantially longer processing times.

Inception~V3 occupies an intermediate position with an average training time of approximately 71~s, which makes it roughly 3$\times$ slower than ResNet and 2$\times$ faster than VGG. However, the moderate variance ($\mathrm{std}=10.4$~s), combined with the presence of quality artifacts (discussed in Section~4.4), makes this architecture less attractive than training time alone might suggest.

\begin{figure}[H]
    \centering
    \includegraphics[width=\linewidth]{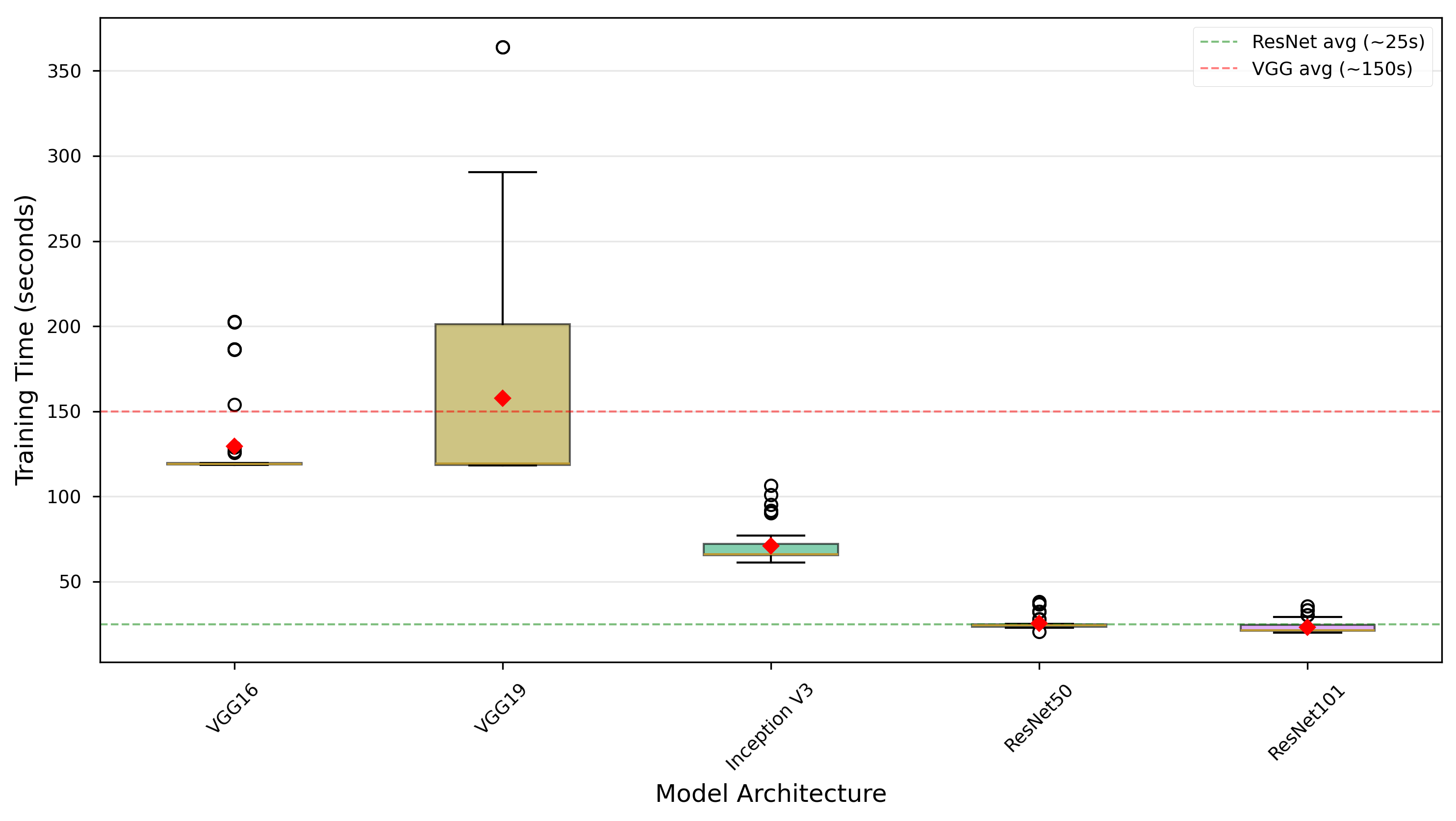}
    \caption{Box plot showing training time distributions across architectures.}
    \label{fig:training_time_boxplot}
\end{figure}

As shown in Fig.~\ref{fig:training_time_boxplot}, the training time distributions highlight clear efficiency differences among architectures. ResNet models (ResNet50: 25.4s, ResNet101: 23.3s) demonstrate compact distributions with minimal variance and few outliers, which indicate consistent optimization. VGG architectures show higher medians (VGG16: 129.5s, VGG19: 157.7s) with wide distributions and numerous outliers exceeding 250 seconds. Inception V3 shows intermediate performance (71.0s) with moderate variance. Horizontal reference lines mark typical ResNet (~25s, green) and VGG (~150s, red) performance levels.

\subsubsection{Convergence Dynamics}

\begin{figure}[H]
    \centering
    \includegraphics[width=\linewidth]{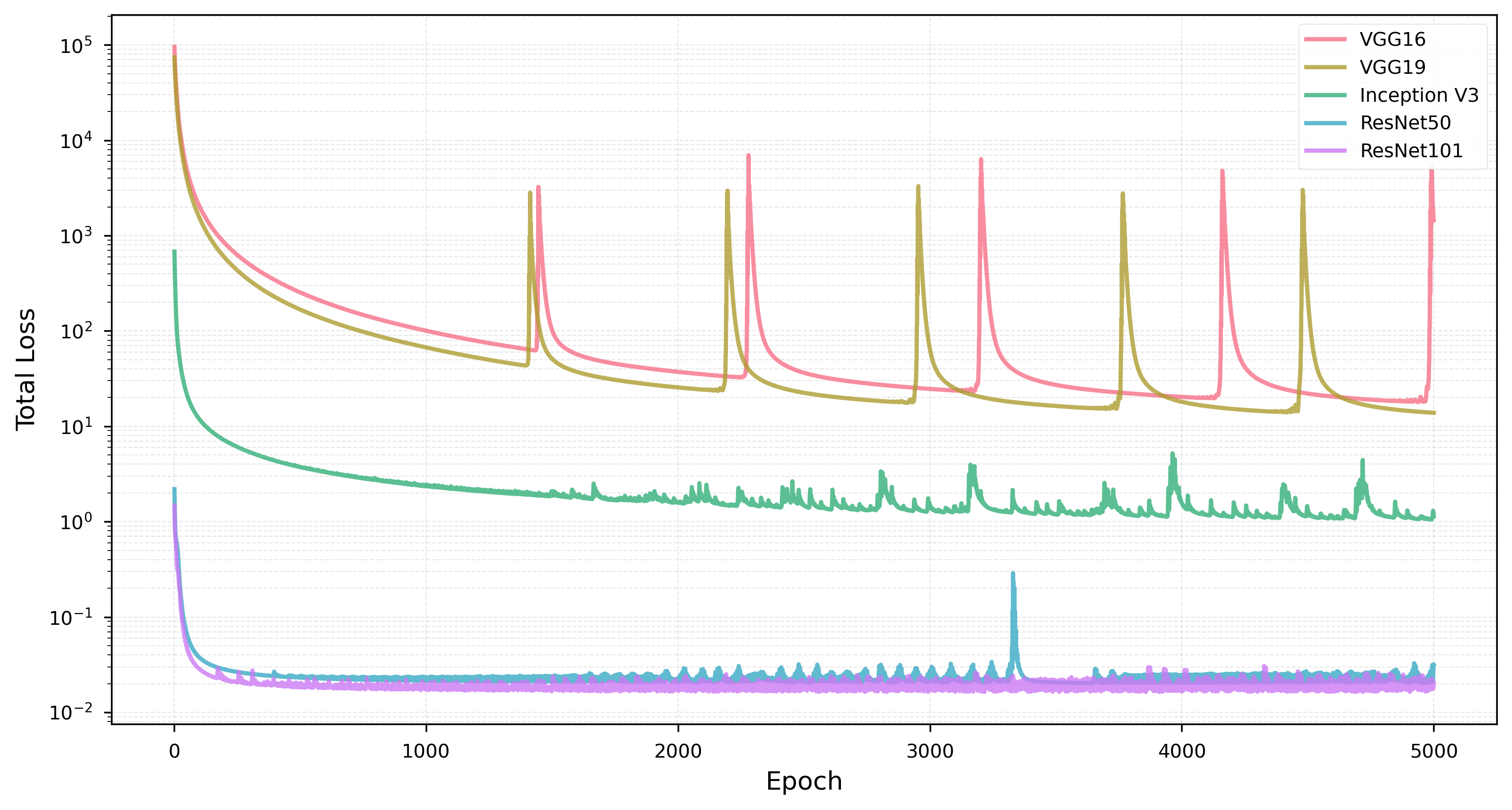}
    \caption{Loss convergence curves comparing architectures.}
    \label{fig:loss_convergence_comparison}
\end{figure}

As shown in Fig.~\ref{fig:loss_convergence_comparison}, the training loss trajectories for representative experiments are presented for all evaluated architectures. Across all architectures, the loss exhibits a clear overall downward trend over 5000 epochs, which indicates successful optimization of the style transfer objective.

ResNet-based models show smooth and rapid convergence, reaching a stable plateau at earlier stages of training, which suggests stable gradient propagation and consistent optimization behavior. VGG architectures display similar long-term convergence trends but with pronounced non-monotonic fluctuations during training. These periodic loss spikes indicate increased sensitivity to optimization dynamics rather than divergence, reflecting less stable convergence behavior compared to residual networks.

Inception~V3 reaches convergence at a rate similar to ResNet architectures, but its training process shows more pronounced fluctuations, indicating reasonably stable yet less smooth optimization behavior. Importantly, differences in absolute loss values across architectures should not be interpreted as direct indicators of output quality, as each architecture computes losses within distinct feature representations.

Although convergence is observed at earlier stages (approximately around 2000 epochs for most models), a fixed training length of 5000 epochs was adopted to ensure full stabilization of the optimization process across all architectures under a unified experimental setting. In practical deployment scenarios, comparable visual outcomes may be achieved with substantially fewer iterations, particularly for architectures exhibiting early convergence behavior.

\section{Discussion}

\subsection{Interpretation of Findings}

Our evaluation reveals that modern CNN architectures achieve broadly similar quality for batik style transfer, with most practical differences emerging in training efficiency, alongside largely comparable output quality. This finding has important implications for both research and practice.

The non-significant ANOVA result ($F = 0.37$, $p = 0.83$) provides strong statistical evidence that architectural choice explains minimal variance in SSIM performance (less than 1\%) based on effect size analysis. This contrasts with some prior NST research this suggests that substantial architectural advantages, but aligns with recent work which focuses on the importance of hyperparameter tuning and training procedures over architecture selection. The convergence of performance across diverse architectures (VGG's sequential design, ResNet's residual connections, Inception's multi-scale processing) suggests that the batik style transfer task may be relatively architecture-agnostic, with optimization dynamics and hyperparameter configuration playing more decisive roles.

ResNet's training efficiency advantage (5--6$\times$ faster than VGG) represents the most actionable finding from our study. Our computational analysis reveals the theoretical foundation for this performance gap: ResNet architectures require only 0.63 GFLOPs per forward pass compared to VGG's 10.12 GFLOPs---a 16$\times$ difference in computational cost. This disparity arises from fundamental architectural differences: ResNet's residual connections enable efficient gradient flow through skip connections, which reduce the number of operations required per inference while maintaining representational capacity through increased depth. In contrast, VGG's sequential convolutional architecture necessitates processing through every layer without shortcuts, yielding in substantially higher computational overhead despite having fewer parameters (14.71--20.02M vs. 23.51--42.50M for ResNet).

The strong correlation between FLOPs and observed training time validates our empirical measurements and demonstrates that ResNet's efficiency advantage is not merely an artifact of implementation or hardware configuration, but rather an inherent property of the architecture. For applications requiring processing of large image batches or rapid iteration during creative workflows, this efficiency difference substantially impacts feasibility and user experience. A system handling 1000 image pairs completes the task in about 7 hours using ResNet, whereas VGG requires 36–44 hours, indicating a gap that has major implications for deployment.

Importantly, this speed advantage comes without sacrificing output quality, which make ResNet architectures particularly attractive for production environments where both quality and throughput matter. The modest GPU memory requirements across all architectures (0.12--0.20 GB peak VRAM) indicate that memory is not a limiting factor, further emphasizing computational efficiency as the primary differentiator for practical deployment.

Inception V3's intermediate position (4.08 GFLOPs, 3$\times$ slower than ResNet) might suggest a reasonable compromise, but the presence of quality artifacts in certain style transfers (discussed in Section~4.4) undermines this potential advantage. The computational savings relative to VGG do not offset the quality concerns, particularly when ResNet offers both superior efficiency and consistent quality.

\subsection{Practical Recommendations}

Based on our qualitative and quantitative analyses, we provide simplified practical recommendations for selecting neural style transfer architectures in batik motif generation. When strong and visually dominant style transfer is the primary objective and longer training time is acceptable, VGG-based models are preferable, as they tend to produce more pronounced and painterly stylistic effects.

For applications where computational efficiency and faster convergence are prioritized, ResNet architectures offer a practical solution. While the resulting style intensity is generally more restrained, ResNet-based models preserve content structure more effectively and achieve stable results with reduced training cost.

In scenarios requiring a balance between stylistic expressiveness and efficiency, Inception V3 may be considered. However, practitioners should be aware that this balance comes at the expense of increased sensitivity to noise and reduced structural stability in certain cases.

The choice of architecture should be guided by the desired trade-off between stylistic strength, training efficiency, and visual stability, rather than minor differences in quantitative evaluation metrics.

\subsection{Limitations}
Several limitations should be acknowledged when interpreting our findings. Our evaluation focused exclusively on Indonesian batik motifs, which exhibit distinctive geometric and organic patterns characteristic of this traditional art form. While this specificity enabled controlled comparison across architectures, generalization to other artistic styles, such as Western impressionism, abstract expressionism, or contemporary digital art which requires independent validation. The visual characteristics and texture complexity of batik may elicit different architectural responses compared to other artistic domains.

The hardware constraints of our experimental setup may influence the observed performance patterns. All experiments were conducted on a GTX 1650 GPU with 4GB VRAM, which shows mid-range consumer hardware. Relative training time advantages and memory efficiency patterns may differ on high-end GPUs with larger memory capacity and higher computational throughput. Additionally, our findings are specific to the Gatys et al. optimization-based NST framework. Feed-forward neural style transfer networks, which employ different optimization strategies and architectural designs, may exhibit different performance characteristics and architectural sensitivities.

Finally, while our sample size of 49 image pairs per model (245 total experiments) provided adequate statistical power for detecting medium-to-large effect sizes, larger datasets spanning more diverse content-style combinations would strengthen the generalizability of our conclusions and enable detection of smaller architectural differences if they exist.

\subsection{Future Directions}
Several promising avenues for future research emerge from our findings. Hybrid architectural approaches that combine ResNet's computational efficiency with VGG's distinctive artistic characteristics represent an intriguing direction. Such architectures might employ ResNet blocks for rapid feature extraction while incorporating VGG-style layers for final stylization, potentially achieving both speed and aesthetic quality advantages.

The integration of attention mechanisms offers another compelling research direction. Spatially-selective style transfer, guided by attention maps that identify salient content regions and appropriate style patterns, could enable more nuanced and contextually-appropriate stylization. This approach might address current limitations in handling complex scenes with multiple distinct objects or regions requiring different stylization intensities.

Transformer-based architectures, which have demonstrated remarkable success in various computer vision tasks, warrant investigation for neural style transfer applications. Their ability to capture long-range dependencies through self-attention mechanisms may prove particularly valuable for batik patterns, which often exhibit global symmetries and repeating motifs across large spatial scales. The computational cost of transformers remains a consideration, but architectural innovations continue to improve their efficiency.

Adaptive hyperparameter optimization based on content complexity represents a practical research direction with immediate applicability. Current NST implementations typically employ fixed hyperparameters across all images, but content-dependent adjustment of learning rates, style weights, and iteration counts could improve both efficiency and quality. Machine learning approaches for predicting optimal hyperparameters from content-style pair characteristics could automate this adaptation.

In addition, large-scale subjective evaluation studies would provide valuable validation of metric-based findings. While quantitative metrics such as SSIM and LPIPS correlate with perceptual quality, human aesthetic preferences, particularly in artistic contexts may diverge from metric rankings. Systematic user studies comparing architectural outputs across diverse viewer populations would complement our quantitative analysis and reveal subjective quality dimensions not captured by current metrics.
\section{Conclusion}
This study presented a systematic comparative evaluation of five CNN architectures (VGG16, VGG19, Inception V3, ResNet50, and ResNet101) specifically for the generative preservation of Indonesian batik motifs. By conducting 245 controlled experiments and rigorous statistical analyses, we challenged the prevailing methodological reliance on VGG-based networks in neural style transfer (NST).

Our findings lead to three primary conclusions. First, \textbf{architectural efficiency is the definitive differentiator, rather than output quality alone}. While statistical analysis (ANOVA, $p = 0.83$) and perceptual metrics (SSIM, LPIPS) revealed no significant quality differences among the models, ResNet-based architectures demonstrated a clear computational advantage. ResNet50 and ResNet101 achieved convergence approximately 5--6$\times$ faster than the industry-standard VGG models. Our computational analysis reveals the theoretical foundation for this performance gap: ResNet requires only 0.63 GFLOPs per forward pass compared to VGG's 10.12 GFLOPs---a 16$\times$ difference in computational cost. This efficiency stems from ResNet's residual connection architecture, which enables gradient flow through skip connections while VGG's sequential design necessitates processing through every layer. With modest GPU memory requirements across all architectures (0.12--0.20 GB), computational efficiency rather than memory becomes the primary constraint. This establishes ResNet as a highly effective backbone for resource-constrained industrial applications, such as digital prototyping for batik SMEs, where throughput and iteration speed are critical.

Second, \textbf{although quantitative metrics are similar, qualitative characteristics differ fundamentally}. VGG architectures tend to produce dense, painterly texturization suitable for highly artistic abstractions, whereas ResNet architectures prioritize structural fidelity, preserving the geometric integrity of original batik lines (canting strokes) more effectively. Consequently, architectural choice should be guided by application requirements: VGG favors maximal stylistic expressiveness, ResNet prioritizes structural preservation and efficiency, while Inception occupies an intermediate balance with modest fine-scale noise.

Third, \textbf{our ablation study confirms that hyperparameter tuning offers greater control over visual aesthetics than architectural selection alone}. Adjustments to learning rates and content--style weight ratios yielded more substantial variations in style intensity (SSIM range 0.28--0.50) compared to switching network backbones (SSIM range 0.27--0.30). Accordingly, for practical deployment, we recommend prioritizing ResNet50 as an efficient baseline architecture, with stylistic refinement achieved through targeted hyperparameter optimization tailored to specific cultural motifs.

Future work should explore hybrid architectures that combine the computational efficiency of ResNet with the rich textural representations of VGG, as well as the integration of attention mechanisms to better handle complex multi-object batik compositions. Additionally, investigating methods to reduce the 16$\times$ FLOPs gap between VGG and ResNet through architectural innovations could enable VGG-quality texturization at ResNet-level efficiency. Ultimately, this research provides a validated technical framework for leveraging generative AI in the safeguarding, adaptation, and sustainable evolution of intangible cultural heritage.

\bibliographystyle{plain}
\bibliography{references}

\end{document}